\documentclass[sigconf]{acmart}
\copyrightyear{2024}
\acmYear{2024}
\setcopyright{acmlicensed}
\acmConference[CIKM '24] {Proceedings of the 33rd ACM International Conference on Information and Knowledge Management}{October 21--25, 2024}{Boise, ID, USA.}
\acmBooktitle{Proceedings of the 33rd ACM International Conference on Information and Knowledge Management (CIKM '24), October 21--25, 2024, Boise, ID, USA}
\acmISBN{979-8-4007-0436-9/24/10}
\acmDOI{10.1145/3627673.3679521}
\usepackage{enumitem}
\usepackage{bm}
\usepackage{multirow}
\usepackage{multicol}
\usepackage{amsfonts}
\usepackage{graphicx}
\usepackage{tabularx}
\usepackage{subfigure}
\usepackage{calc}
\newcommand{\modelname}{\texttt{PRISM}}

\AtBeginDocument{%
  \providecommand\BibTeX{{%
    \normalfont B\kern-0.5em{\scshape i\kern-0.25em b}\kern-0.8em\TeX}}}

\begin{document}

\title{\modelname{}: Mitigating EHR Data Sparsity via Learning from Missing Feature Calibrated Prototype Patient Representations}

\settopmatter{authorsperrow=4}
\author{Yinghao Zhu}
\orcid{0000-0002-2640-6477}
\affiliation{%
  \institution{Beihang University \\ Peking University}
  \city{Beijing}
  \country{China}
}
\email{yhzhu99@gmail.com}

\author{Zixiang Wang}
\orcid{0009-0000-1257-9580}
\affiliation{%
  \institution{Peking University}
  \city{Beijing}
  \country{China}
}
\email{wangzx@stu.pku.edu.cn}

\author{Long He}
\orcid{0009-0002-3663-3718}
\affiliation{%
  \institution{Tsinghua University}
  \city{Beijing}
  \country{China}
}
\email{longhe0820@gmail.com}

\author{Shiyun Xie}
\orcid{0000-0001-5921-4060}
\affiliation{%
  \institution{Beihang University}
  \city{Beijing}
  \country{China}
}
\email{xieshiyun@buaa.edu.cn}

\author{Xiaochen Zheng}
\orcid{0009-0007-9714-2193}
\affiliation{%
  \institution{ETH Zürich}
  \city{Zürich}
  \country{Switzerland}
}
\email{xzheng@ethz.ch}

\author{Liantao Ma}
\authornote{Corresponding author.}
\orcid{0000-0001-5233-0624}
\affiliation{%
  \institution{Peking University}
  \city{Beijing}
  \country{China}
}
\email{malt@pku.edu.cn}

\author{Chengwei Pan}
\authornotemark[1]
\orcid{0000-0003-0497-7903}
\affiliation{%
  \institution{Beihang University \\ Zhongguancun Laboratory}
  \city{Beijing}
  \country{China}
}
\email{pancw@buaa.edu.cn}

\renewcommand{\shortauthors}{Yinghao Zhu et al.}
\renewcommand{\shorttitle}{\modelname{}: An EHR Data Sparsity Mitigation Framework}

\begin{abstract}

Electronic Health Records (EHRs) contain a wealth of patient data; however, the sparsity of EHRs data often presents significant challenges for predictive modeling. Conventional imputation methods inadequately distinguish between real and imputed data, leading to potential inaccuracies of patient representations. To address these issues, we introduce \modelname{}, a framework that indirectly imputes data by leveraging prototype representations of similar patients, thus ensuring compact representations that preserve patient information. \modelname{} also includes a feature confidence learner module, which evaluates the reliability of each feature considering missing statuses. Additionally, \modelname{} introduces a new patient similarity metric that accounts for feature confidence, avoiding over-reliance on imprecise imputed values. Our extensive experiments on the MIMIC-III, MIMIC-IV, PhysioNet Challenge 2012, eICU datasets demonstrate \modelname{}'s superior performance in predicting in-hospital mortality and 30-day readmission tasks, showcasing its effectiveness in handling EHR data sparsity. For the sake of reproducibility and further research, we have publicly released the code at \url{https://github.com/yhzhu99/PRISM}.

\end{abstract}

\ccsdesc[500]{Applied computing~Health informatics}
\ccsdesc[500]{Information systems~Data mining}

\keywords{electronic health record; data mining; deep learning}

\maketitle

\section{Introduction}\label{sec:introduction}

Electronic Health Records (EHR) have become indispensable in modern healthcare, offering a rich source of data that chronicles a patient's medical history. Over recent years, machine learning techniques have gained significant attention for their ability to leverage time-series EHR data, which represented as temporal sequences of high-dimensional clinical variables~\cite{choi2016retain}, can significantly inform and enhance clinical decision-making. Such applications range from predicting the survival risk of patients~\cite{ma2020concare,zhang2021grasp,liao2024learnable} to forecasting early mortality outcomes~\cite{yan2020interpretable,ma2022safari,gao2024comprehensive,zhu2023m3fair}.

Working with time-series EHR data presents challenges due to its inherent sparsity. Factors such as data corruption~\cite{chen2020hgmf}, expensive examinations~\cite{ford2000non}, and safety considerations~\cite{zhou2020hi} result in missing observations; for instance, not all indicators are captured during every patient visit~\cite{che2018recurrent}. Imputed values, while necessary, are not genuine reflections of a patient's condition and can introduce noise, diminishing model accuracy~\cite{zhang2022m3care}. Given that most machine learning models cannot process NaN (Not a Number) inputs, this sparsity necessitates imputation, complicating EHR predictive modeling. While most existing works tackling EHR sparsity have tried to perform the imputation task directly on raw data, based on modeling the health status trajectory of the whole training set, this approach is similar to that of matrix completion methods, such as MICE~\cite{van2011mice}, non-negative matrix factorization (NMF)~\cite{wang2012nonnegative}, and compressed sensing~\cite{lopes2013estimating}. However, they fail to capture the temporal interactions of longitudinal EHR data for each patient. Also, previous EHR-specific models with strategies of recalibrating patient representations based on attentive feature importance~\cite{ma2020adacare,ma2023aicare} exhibit an issue of inadvertently prioritizing imputed features, potentially introducing inaccuracies~\cite{liu2023handling}. Thus, addressing the sparsity in time-series EHR data requires a focus on capturing relevant feature representations across visits within a patient or among patients that are essential for predictive purposes. The primary objective is to discern and highlight crucial features while diminishing the impact of irrelevant, redundant, or missing ones.

Intuitively, incorporating knowledge from similar patients as an indirect method of imputation has the potential to enhance patient representations in the context of sparse EHR data. Such a knowledge-driven approach mirrors the real-world clinical reasoning processes, harnessing patterns observed in related patient cases~\cite{fenglong2018patsimilarity}. Regarding the identifying similar patient process, however, existing models also face a significant limitation: they cannot distinguish between actual and imputed data~\cite{zhang2021grasp,zhang2022m3care}. Consider two patients who have the same lab test feature value. For patient A, this value originates from the actual data, while for patient B, it is an imputed value. Current similarity metrics, whether they are L1, L2 distance, cosine similarity~\cite{lee2015patcossim}, handcrafted metrics~\cite{huang2019ehrsim}, or learning-based methods~\cite{fenglong2018patsimilarity}, interpret these values identically. This results in a potentially misleading perception of similarity.

Given these insights, we are confronted with a pressing challenge: \textbf{How can we effectively mitigate the sparsity issue in EHR data caused by missing recorded features while ensuring a compact patient representation that preserves patient information?~\cite{si2021deep}}

To address this, we introduce \modelname{} that leverages prototype similar patients representations at the hidden state space. Unlike traditional direct imputation methods that imputes values based only raw data, \modelname{} learns refined patient representations according to prediction targets, serving as a more effective imputation strategy and thus mitigating the EHR data sparsity issue. Central to our approach is the feature-missing-aware calibration process in the proposed feature confidence learner module. It evaluates the reliability of each feature, considering its absence, the time since the last recorded visit, and the overall rate of missing data in the dataset. By emphasizing feature confidence, our newly designed patient similarity measure provides evaluations based not just on raw data values, but also on the varying confidence levels of each feature. 

In healthcare, the absence of data can severely compromise confidence in a prognosis. Addressing and understanding the challenges of these missing features is of utmost importance. In light of this, \modelname{} seeks to bridge this gap. Our primary contributions are:
\begin{itemize}[leftmargin=*]
    \item \textbf{Methodologically,} we propose \modelname{}, a framework for learning prototype representations of similar patients, designed to mitigate EHR data sparsity. We design the feature confidence learner that evaluates and calibrates the reliability of each feature by examining its absence and associated confidence level. We also introduce the confidence-aware prototype patient learner with enhanced patient similarity measures that differentiates between varying feature confidence levels. Compared to existing SOTA baselines, i.e. GRASP~\cite{zhang2021grasp} and M3Care~\cite{zhang2022m3care}, \modelname{} provides a refined feature calibration, and further missing-aware similarity measure helps to identify more related patient representation, with missing feature status elaborately taking into account.
    \item \textbf{Experimentally}, comprehensive experiments on four real-world datasets, MIMIC-III, MIMIC-IV, Challenge-2012, and eICU, focusing on in-hospital mortality and 30-day readmission prediction tasks, reveal that \modelname{} significantly improves the quality of patient representations against EHR data sparsity. \modelname{} outperforms the best-performing baseline model with relative improvements of 6.40\%, 2.78\%, 1.51\% and 11.01\% in mortality on AUPRC for four datasets. In terms of readmission task, \modelname{} obtains relative improvements of 1.38\% and 1.63\% on AUPRC for MIMIC-III and MIMIC-IV, respectively. 
    Further ablation studies and detailed experimental analysis underline \modelname{}'s effectiveness, robustness, adaptability, and efficiency.
\end{itemize}

\section{Related Work}\label{sec:related_work}

In the realm of EHR data analysis, irregular sampling often leads to significant data sparsity, presenting substantial challenges in modeling. Previous methods on sparse EHR data predominantly fall under two categories: direct imputation in the raw data space and indirect imputation within the feature representation space.

\subsection{Direct Imputation}
Direct imputation methods aim to estimate missing features or incorporate missing information directly. Traditional matrix imputation techniques, such as MICE~\cite{van2011mice}, non-negative matrix factorization (NMF)~\cite{wang2012nonnegative}, compressed sensing~\cite{lopes2013estimating}, or naive zero, mean, or median imputation, rely on similar rows or columns to fill in missing data. However, these methods often operate under the assumption that patient visits are independent and features are missing at random. GRU-D~\cite{che2018recurrent} adopts a more targeted approach by introducing missing statuses into the GRU network. By utilizing time interval and missing mask information, GRU-D treats missing data as ``Informative Missing''. Extending these capabilities, ConCare~\cite{ma2020concare} and AICare~\cite{ma2023aicare} first apply directly imputed EHR data, then incorporate multi-head self-attention mechanisms to refine feature embeddings. This ensures contextual relevance across diverse healthcare situations, regardless of data completeness.

\subsection{Indirect Imputation}
As demonstrated in GRASP~\cite{zhang2021grasp} and M3Care~\cite{zhang2022m3care}, they emphasize the use of similar patient representations to derive meaningful information with the insight that the information observed from similar patients can be utilized as guidance for the current patient's prognosis~\cite{zhang2021grasp}. However, accurately measuring patient similarity is intrinsically challenging, especially when features might be imputed with potentially misleading information. Many traditional works, such as ~\cite{lee2015patcossim} and ~\cite{huang2019ehrsim}, have resorted to fixed formulas like cosine similarity and Euclidean distance to gauge patient similarity. While these methods are straightforward, they often suffer from scalability and performance limitations. A more dynamic approach is seen in ~\cite{fenglong2018patsimilarity}, which adopts metric learning with triplet loss. This technique focuses on learning the relative distances between patients, where distances have an inverse correlation with similarity scores.

However, a shared oversight across the aforementioned methods is the underestimation of the impact of missing features. This is evident both during the recalibration of features and when assessing patient similarities, as highlighted in the introduction and essential to tackle the EHR sparsity issue.

\section{Problem Formulation}\label{sec:problem_formulation}

\subsection{EHR Datasets Formulation} EHR datasets consist of a sequence of dynamic and static information for each patient. Assuming that there are $F$ features in total, $D$ dynamic features (e.g., lab tests and vital signs) and $S$ static features (e.g., sex and age), where $F = D + S$, at every clinical visit $t$. The features recorded at visit $t$ can be denoted as $\bm{x}_t \in \mathbb{R}^{F}, t=1,2,\cdots, T$, with total $T$ visits. The dynamic feature information can be formulated as a 2-dimensional matrix $\bm{d} \in \mathbb{R}^{T\times D}$, along with static information denoted as 1-dimensional matrix $\bm{s} \in \mathbb{R}^{S}$. In addition, to differentiate between categorical and numerical variables within dynamic features, we employ one-hot encoding for categorical variables. Due to the inherent sparsity of EHR data, we incorporate feature missingness as inputs. At a global view, we define the missing representation, denoted as \(\rho_i\), to be the presence rate of the \(i\)-th feature within the entire dataset. From a local view, the missing representation, \(\tau_{i,t}\), signifies the time interval since the last recorded visit that contains the \(i\)-th feature up to the \(t\)-th visit.

\subsection{Predictive Objective Formulation} Prediction objective is presented as a binary classification task. Given each patient's EHR data $\bm{X} = [\bm{x_1}, \bm{x_2}, \cdots, \bm{x_T}]^{\top} \in \mathbb{R}^{T \times F}$ and feature missing status $\{\rho, \tau\}$ as input, where each $\bm{x_t}$ consists of dynamic features and static features representation, the model attempts to predict the specific clinical outcome, denoted as $y$. The objective is formulated as $\hat{y} = \text{Model}(\bm{X}, \{\rho, \tau\} )$. For the in-hospital mortality prediction task, the goal is to predict the discharge status ($0$ for alive, $1$ for deceased) based on the initial 48-hour window of an ICU stay. Similarly, the 30-day readmission task predicts if a patient will be readmitted in 30 days ($0$ for no readmission, $1$ for readmission).

\subsection{Notation Table}
Table~\ref{tab:notation} contains notation symbols and their descriptions used in the paper.

\begin{table}[!ht]
\footnotesize
\caption{\textit{Notations symbols and their descriptions}}
\centering
\label{tab:notation}
\begin{tabular}{c|p{5.5cm}}
    \toprule
    \textbf{Notations} & \multicolumn{1}{c}{\textbf{Descriptions}} \\
\midrule
    $N$ & Number of patient samples\\
    $T$ & Number of visits for a certain patient\\
    $D$ & Number of dynamic features\\
    $S$ & Number of static features\\
    $F$ & Number of features, $F = S+D$\\
    $ \bm{d_{it}} \in \mathbb{R}^{m}$ & The \( i \)-th feature at the \( t \)-th visit, where \( m \) is either the number of categories (for one-hot encoding) or 1 (for numerical lab tests) \\ 
    \( \bm{X} \in \mathbb{R}^{T \times F} \) & Clinical visit matrix of a single patient, consisting of $T$ visits \\
    $ \bm{s}, \bm{d} $  & Static and dynamic feature vector of a patient\\
    $ \bm{y}, \bm{\hat{y}}$ & Ground truth labels and prediction results\\
\midrule
    $\bm{h_{i}} \in \mathbb{R}^{T \times f}, \bm{h} $ & Representation of $i$-th feature learned by GRU (for dynamic features) or MLP (for static features), stacked to form the representation matrix \( \bm{h} \), \(f\) is each feature's embedding dimension \\
    \( \rho_{i} \) & Feature presence rate of feature $i$ in training set \\
    \( \tau_{i, t} \) & Time interval from the last recorded visit of $i$-th feature at $t$-th visit \\
    \( \bm{C} \in \mathbb{R}^{T \times F} \) & Learned feature confidence matrix of a patient \\
    \( \bm{z_{i}} \) & \( \bm{z_{i}} \) is the learned representation of the \(i\)-th patient after the feature calibration layer\\
    \( \bm{\alpha}, \bm{\alpha^*} \) & Learned attention weights and calibrated attention weights after the feature calibration layer \\
    \( \phi(\cdot, \cdot) \) & Patient similarity measure function \\
    \( \bm{A} = (a_{i,j}) \) & Adjacency matrix of patients, composed of similarity score between \(i\)-th and \(j\)-th patient \\
    \( \bm{\mathcal{G}_{k}} \) &  Learned prototype patient representation of the $k$-th group \\
    \( \bm{z^{*}_{i}} \) & Learned representation of the \(i\)-th patient after representation fusion layer \\
\midrule
    $\boldsymbol{W}_{\square}$ & Parameter matrices of linear layers. Footnote $\square$ denotes the name of the layer \\
    \( K \) & Number of similar patient groups \\
    \bottomrule
\end{tabular}
\end{table}

\section{Methodology}\label{sec:methods}

\subsection{Overview}

\begin{figure*}[!t]
    \centering
    \includegraphics[width=1.0\linewidth]{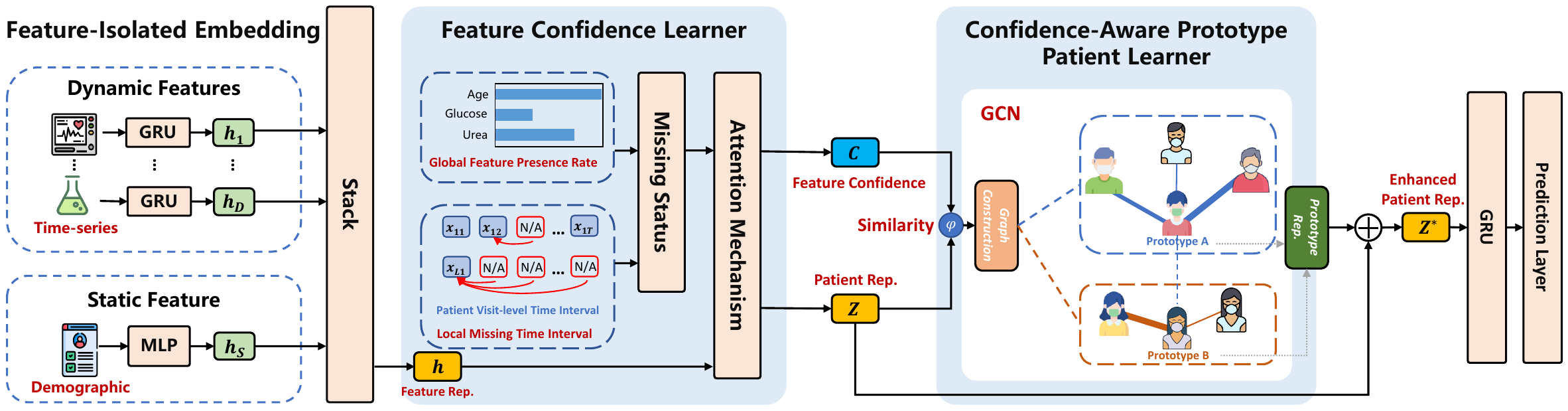}
    \caption{\textit{Overall model architecture of our proposed method \modelname{}.} ``Rep.'' means ``Representation''.}
    \label{fig:pipeline}
\end{figure*}

Figure~\ref{fig:pipeline} shows the overall pipeline of \modelname{}. It consists of three main sub-modules below.

\begin{itemize}[leftmargin=*]
    \item \textbf{Feature-Isolated Embedding Module} applies GRU and MLP backbone separately to dynamic features and static features. Each dynamic feature learns historical representations over multiple time steps. To align with the original attribute information of each feature, the features are learned in isolation from each other.
    \item \textbf{Feature Confidence Learner} improves self-attention model by introducing the feature missing status (the global dataset-level and local patient-level missing representations of the features), collaboratively learning the confidence level of the features and the confidence-calibrated feature importance.
    \item \textbf{Confidence-Aware Prototype Patient Learner} improves the measure of patient similarity based on patient representation and the confidence level of features learned from feature missing status. It then applies the graph neural network to learn prototype patients. Finally, the patient's own representation is fused with the prototype patient representation adaptively, further enhancing the hidden state representation of the patient that is affected by missing data. A two-layer GRU network is then utilized to generate a compact patient health representation, followed by a single-layer MLP network to conduct downstream specific prediction tasks.
\end{itemize}

\subsection{Feature-Isolated Embedding Module}

In this module, static and dynamic features are learned individually via Multilayer Perceptron (MLP) and Gated Recurrent Unit (GRU) networks, yielding feature representations of unified dimensions $f$.

\subsubsection{\textbf{Static Features Embedding}}
\mbox{}\\
Static features remain constant at each visit. Hence, we opt for a single-layer MLP for simplicity to map each static feature into the feature dimensions $f$:
\begin{equation}
\begin{array}{cc}
\bm{h_{s_i}}=\mathrm{MLP}_i(\bm{s_i}), & i=1,2,\cdots,S
\end{array}
\end{equation}
where $\bm{s_i}$ is the $i$-th static feature. We employ $S$ distinct-parameter MLPs for feature mappings.

\subsubsection{\textbf{Dynamic Features Embedding}}
\mbox{}\\
To ensure that each feature's individual statistics, e.g. missing status can be incorporated with the corresponding feature, we adopt multi-channel GRU structure to avoid feature interaction at this stage. Each feature is embed with an isolated GRU, a time-series model that has a proven track record of consistent performance in EHR modeling~\cite{gao2024comprehensive}:
\begin{equation}
\begin{array}{cc}
\bm{h_{d_i}}=\mathrm{GRU}_i(\bm{d_i}), & i=1,2,\cdots,D
\end{array}
\end{equation}
where $\mathrm{GRU}_i$ represents the GRU network applied to the $i$-th dynamic feature $\bm{d_i} \in \mathbb{R}^{T \times m}$. Furthermore, $\bm{h_{d_i}} \in \mathbb{R}^{T \times f}$ signifies the embedding of the $i$-th dynamic feature. The in-channel of GRU is the feature recorded dimension $m$, and the out-channel is the unified $f$.

Then we employ a stack operation to integrate information from both static and dynamic features. This necessitates initially replicating the static features embeddings to each time visits: $\bm{h_{s_i}} \in \mathbb{R}^{f} \rightarrow \bm{h'_{s_i}} \in \mathbb{R}^{T \times f}$. The stack operation is represented as follows:
\begin{equation}
    \bm{h}=\mathrm{stack}(\bm{h'_{s_1}},\bm{h'_{s_2}},\cdots,\bm{h'_{s_S}}, \bm{h_{d_1}},\cdots,\bm{h_{d_D}})
\end{equation}
where $\bm{h} \in \mathbb{R}^{F \times T \times f}$ represents the overall embeddings of features.

\subsection{Feature Confidence Learner}

Existing models adopt various ways to enhance patient representations to mitigate the noise introduced by processing sparse EHR data. However, these models often use imputed data and ignore the impact of feature missing status, thus reducing the credibility of the learned hidden representation. We design a measurement called ``feature confidence'', which represents the reliability of the input feature values for each patient and each visit. In addition, we have incorporated this measure into the self-attention mechanism as a recalibration module to elevate low-confidence features' attention.

\subsubsection{\textbf{Feature Missing Status Representation}}
\mbox{}\\
We introduce \(\rho\) and \(\tau\) to record feature missing status. Global missing representation $\rho_i$ represents the presence rate of the $i$-th feature in the original dataset:
\begin{equation}
\rho_i = \frac{\text{total observations of}\ i\text{-th feature}}{\text{total visits of all patients}}
\end{equation}
For example, if the dataset collects a total of 100 data records during the visits of all patients, but certain feature is only recorded two times, then the $\rho$ for this feature is $\frac{2}{100}=0.02$. Local missing representation $\tau_{i, t}$ represents the time interval since the last record of this feature at the current visit. There are two special cases: case 1) If the feature is recorded at the current visit, it is marked as 0; case 2) if the feature has never been recorded before the current visit, it is marked as infinity:
\begin{equation}
    \tau_{i, t} = 
    \begin{cases} 
    0 & \text{if case 1)} \\
    \infty & \text{if case 2)} \\
    t - t^{*} & \text{otherwise}
    \end{cases}
\end{equation}
where $t^*$ is the time of the last record of the $i$-th feature.

\subsubsection{\textbf{Missing-Aware Self-Attention}}
\mbox{}\\
To calculate the feature confidence, we comprehensively consider the missing feature status in the dataset, including the global missing representation \(\rho\) and local missing representation \(\tau\), and integrate them into a self-attention mechanism module.

First, the $Query$ vector is computed from the hidden representation of the last time step $T$, while the $Key$ and $Value$ vectors are computed from the hidden representations of all time steps:
\begin{equation}
    {q}_{i,T} = \boldsymbol{W}_{i}^{q}\cdot {h}_{i,T}
\end{equation}
\begin{equation}
    {k}_{i,t} = \boldsymbol{W}_{i}^{k}\cdot {h}_{i,t}
\end{equation}
\begin{equation}
    {v}_{i,t} = \boldsymbol{W}_{i}^{v}\cdot {h}_{i,t}
\end{equation}
where ${q}_{i,T}$, ${k}_{i,t}$, ${v}_{i,t}$ are the $Query$, $Key$, $Value$ vectors respectively, and $\boldsymbol{W}_{i}^{q}$, $\boldsymbol{W}_{i}^{k}$, $\boldsymbol{W}_{i}^{v}$ are the corresponding projection matrices. Following this, we compute the attention weights as follows:
\begin{equation}
\bm{{\alpha}}_{i,*,t}={softmax}(\frac{{q}_{i,T}\bm{{k}}_{*,t}^\top}{\sqrt{d_k}}) 
\end{equation}

Subsequently, the feature confidence learner takes into account both feature missing status and attention weights to compute the feature confidence, which serves as an uncertainty reference when identifying similar patients in subsequent steps:
\begin{equation}
C_{i,t} = 
\begin{cases} 
\tanh\left(\frac{\alpha_{i,t}}{\omega_{i,t}}\right) & \text{if } \tau_{i,t}\neq \infty \\
\beta\cdot\rho_{i} & \text{if } \tau_{i,t}=\infty 
\end{cases}
\end{equation}
where $\omega_{i,t}$ and $\alpha_{i,t}$ are defined as:
\begin{equation}
\omega_{i,t} = {\gamma_{i}\cdot\log(e+(1-\alpha_{i,t})\cdot\tau_{i,t})}
\end{equation}
\begin{equation}
\alpha_{i,t} = AvgPool(\bm{\alpha}_{i,*,t})
\end{equation}

Here, the global missing parameter $\beta$ is a learnable parameter for global missing representation and the time-decay ratio $\gamma_i$ is a feature-specific learnable parameter to reflect the influence of local missing representation as time interval increases. The calculation of feature confidence is divided into two scenarios: 
\begin{itemize}[leftmargin=*]
    \item When the feature has been examined in previous visits, the authentic values of the same patient are often used to complete features. However, the feature confidence of imputed values for this feature should significantly diminish when:
    \begin{itemize}
        \item time interval $\tau_{i, t}$ is large. As the time interval increases, the feature confidence will decay sharply.
        \item the time-decay ratio $\gamma_i$ is high. The higher the time-decay ratio, the more severe the decay of the feature confidence level, and only the most recent recorded data matters.
    \end{itemize}
    \item Until the present visit, the examination of this specific feature has not been conducted. In this scenario, the imputed values are derived from other patients and the global missing representation $\rho$ is selected to depict the feature confidence for the current patient.
\end{itemize}

Finally, based on feature confidence, we can obtain the calibrated attention weights $\bm{\alpha^*}$ and further hidden representations $\bm{z}$.

\begin{equation}
    \bm{\alpha^*}=\epsilon\cdot\bm{\alpha}+(1-\epsilon)\cdot\bm{C}
\end{equation}
\begin{equation}
    \bm{z}=\bm{\alpha^*}\bm{V}
\end{equation}
where $\epsilon$ is a learnable parameter, $\bm{\alpha}$, $\bm{C}$, $\bm{\alpha^*}$, $\bm{V}$, $\bm{z}$ $\in \mathbb{R}^{T \times F}$ are attention weights, feature confidence, calibrated attention weights, value vector and learned representation of the input patient.

\subsection{Confidence-Aware Prototype Patient Learner}

We incorporate feature confidence status into our confidence-aware module for identifying similar patient cohorts, accounting for the impact of missing features. Inspired by GRASP's~\cite{zhang2021grasp} graph convolutional network (GCN) framework, we further compute and integrate the similarity score within the GCN, considering the individual patient's feature confidence to prioritize the selection of patients most similar to the subject patient.

\subsubsection{\textbf{Confidence-Aware Patient Similarity Measure}}
\mbox{}\\
To find similar patients, we calculate the similarity between the current patient and others using the confidence-aware patient similarity measure, resulting in a similarity score matrix $\bm{A}$. Specifically, the similarity score \( \phi_{i,j}(\bm{z}_i,\bm{z}_j;\bm{C}_i,\bm{C}_j) \) between the $i$-th and $j$-th ($i \neq j$) patients is defined as Equation~\ref{eq:phi}. Note that when $i = j$, indicating the comparison of a patient with themselves, the similarity score is defined in $\phi_{i,j} = 1$.
\begin{equation}
\label{eq:phi}
\phi_{i,j} =
\frac{1}{(1-\zeta) \cdot \psi_{i,j}^{(z)}(\bm{z}_i,\bm{z}_j) + 
\zeta \cdot \psi_{i,j}^{(C)}(\bm{C}_i,\bm{C}_j)}
\end{equation}
where $\psi_{i,j}^{(z)}(\bm{z}_i,\bm{z}_j)$ measures the similarity of the patients' representations,  $\psi_{i,j}^{(C)}(\bm{C}_i,\bm{C}_j)$ measures the confidence level of the two patients in their respective feature representations, and $\zeta$ serves as learnable weight to balance the two. They are defined as follows respectively:
\begin{equation}
\psi_{i,j}^{(z)}(\bm{z}_i,\bm{z}_j) = \frac{1}{F} \|\bm{z}_i - \bm{z}_j\|_2^2
\end{equation}
\begin{equation}
\psi_{i,j}^{(C)}(\bm{C}_i,\bm{C}_j) = \frac{1}{F} \sum_{k=1}^{F} \exp(1-C_{i,k})\cdot\exp(1-C_{j,k})
\end{equation}

\subsubsection{\textbf{Prototype Patients Cohort Discovery}}
\mbox{}\\
To compute the enhanced representations of similar patients, we design the prototype patients cohort discovery module by incorporating GCN's capability to learn relationships between graph nodes. Initially, we utilize the K-Means clustering algorithm to cluster raw patient representation ($F$ recorded or imputed features) into $K$ groups. Subsequently, we identify the center vectors of $K$ clusters as the initial $K$ prototypes $\bm{\mathcal{G}_k}, k=1,2,\cdots, K$. These prototypes vectors, along with the patient groups, form the nodes of a graph. We then calculate the edge weights for this graph using a predefined similarity measure, resulting in an adjacency matrix $\bm{A}$. As GCN learns across epochs, the graph structure dynamically evolves. Note that the feature confidence $\mathbf{C}$ of prototype patients is initially set to $1$ for each feature. During the training phase, both $\bm{\mathcal{G}}$ and the corresponding $\mathbf{C}$ are adaptively adjusted. Consequently, the most similar and representative prototypes $\bm{\mathcal{G}}$ are identified, and samples within the clusters become more similar to each other. The process is illustrated as:
\begin{equation}
\bm{\mathcal{G}^*}={\rm MLP(GCN}(
{\mathrm{concat}(\bm{z}, \bm{\mathcal{G}})}, \bm{A}))
\end{equation}
where $\bm{\mathcal{G}^*}$ is the updated prototype representation by GCN and MLP.

\subsubsection{\textbf{Prototype Representation Fusion}}
\mbox{}\\
Currently, there are two learned hidden representations, one is $\bm{z}$ obtained through the missing-aware self-attention module, and the other is $\bm{\mathcal{G}}$ obtained through the prototype similar patient cohort discovery module. Thus, the patient representation is fused as:
\begin{equation}
\bm{z_{i}^{*}}=\eta\cdot\bm{\mathcal{G}_i}
+(1-\eta)\cdot\bm{z_{i}}
\end{equation}
where $\eta$ is a learnable weight parameter, $\bm{\mathcal{G}_i}$ is the corresponding prototype of the $i$-th sample.

\subsection{Prediction Layers}

Finally, the fused representation $\bm{z^*}$ is expected to predict downstream tasks. We sequentially pass $\bm{z^*}$ through two-layer GRU and a single-layer MLP network to obtain the final prediction results $\hat{y}$:
\begin{equation}
\begin{aligned}
    \hat{\bm{y}}={\rm MLP(GRU}(\bm{z^*}))
\end{aligned}
\end{equation}

The BCE Loss is selected as the loss function for the binary mortality outcome prediction task:
\begin{equation}
    \begin{aligned}
    \mathcal{L}(\hat{y}, y) = -\frac{1}{n}\sum_{i=1}^{n}(y_i \log(\hat{y}_i) + (1 - y_i) \log(1 - \hat{y}_i)) 
    \end{aligned}
\end{equation}
where $n$ is the number of patients within one batch, $\hat{y}\in [0,1]$ is the predicted probability and $y$ is the ground truth.

\section{Experimental Setups}

\subsection{Benchmarked Real-World Datasets}

We employ MIMIC-III, MIMIC-IV, PhysioNet Challenge 2012, and eICU datasets for benchmarking. All 4 datasets are split into 70\% training set, 10\% validation set and 20\% test set with stratified shuffle split strategy based on patients' end-stage mortality outcome. By default, we use the Last Observation Carried Forward (LOCF) imputation method~\cite{woolley2009last}.

\begin{enumerate}[leftmargin=*]
    \item \textbf{MIMIC-III}~\cite{johnson2016mimic} (Medical Information Mart for Intensive Care) is a large, freely-available database comprising information such as demographics, vital sign measurements made at the bedside, laboratory test results, procedures, medications, caregiver notes, imaging reports, and mortality.
    \item \textbf{MIMIC-IV}~\cite{johnson2023mimic} as MIMIC-III dataset's subsequent iteration, stands as an evolved manifestation of the MIMIC-III database, encompassing data updates and partial table reconstructions. We extracted patient EHR data following~\cite{harutyunyan2019multitask}.
    \item \textbf{PhysioNet Challenge 2012}~\cite{silva2012challenge} (Challenge-2012) covers records from 12,000 adult ICU stays. The challenge is designed to promote the development of effective algorithms for predicting in-hospital mortality based on data from the first 48 hours of ICU admission. We utilize 35 lab test features and 5 demographic features in Challenge-2012 dataset.
    \item \textbf{eICU}~\cite{pollard2018eicu} is a large-scale, multi-center ICU database derived from over 200,000 ICU admissions across the United States between 2014 and 2015. 12 lab test features and 2 demographic features are adopted.
\end{enumerate}

The statistics of datasets is in Table~\ref{tab:statistics_datasets}.

\begin{table}[!ht]
        \footnotesize
        \centering
        \caption{\textit{Statistics of datasets after preprocessing.} The proportion demonstrates the percentage of the label with value $1$. $Out.$ denotes Mortality Outcome, $Re.$ denotes Readmission.}
        \label{tab:statistics_datasets}
    \begin{tabular}{c|c|c|c|c}
\toprule
\textbf{Dataset}               & \textbf{MIMIC-III} & \textbf{MIMIC-IV} & \textbf{Challenge-2012} & \textbf{eICU}    \\
\midrule
\# Samples               & 41517     & 56888    & 4000           & 73386   \\
Missing               & 69.87\%   & 74.70\%  & 84.68\%        & 42.61\% \\
$\text{Label}_{Out.}$ & 10.62\%   & 9.55\%   & 13.85\%        & 8.32\%  \\
$\text{Label}_{Re.}$  & 14.74\%   & 13.85\%  & /              & /       \\
\bottomrule
\end{tabular}
\end{table}

\subsection{Evaluation Metrics}

We assess the binary classification performance using AUROC, AUPRC and F1. Here we emphasize AUPRC as the main metric due to it is informative when dealing with highly imbalanced and skewed datasets~\cite{kim2022auprc,davis2006auprc} as shown in our selected datasets.

\subsection{Baseline Models}

We include imputation-based methods, EHR-specific models, and \modelname{} model with reduced modules as baseline models.

\subsubsection{\textbf{Imputation-based Methods}}
\mbox{}\\
We include two imputation-based methods: MICE and GRU-D:
\begin{itemize}[leftmargin=*]
    \item MICE~\cite{van2011mice} addresses missing data in EHR through iterative imputation, with subsequent analysis using an LSTM model~\cite{hochreiter1997lstm}.
    \item GRU-D~\cite{che2018recurrent} incoporates both the last observed and global mean values in the GRU network. Additionally, GRU-D utilizes an exponential decay mechanism to manage the temporal dynamics of missing values.
\end{itemize}

\subsubsection{\textbf{EHR-specific Models}}
\mbox{}\\
Following methods are specifically designed for EHR data and focus on personalized health status embeddings.
\begin{itemize}[leftmargin=*]
    \item RETAIN~\cite{choi2016retain} is a hierarchical attention-based interpretable model. It attends the EHR data in a reverse time order so that recent clinical visits are likely to receive higher attention.
    \item AdaCare~\cite{ma2020adacare} is a GRU-based network that utilizes a multi-scale dilated convolutional module to capture the long and short-term historical variation.
    \item ConCare~\cite{ma2020concare} utilizes multi-channel GRU with a time-aware attention mechanism to extract clinical features and re-encode the clinical information by capturing the interdependencies between features.
    \item GRASP~\cite{zhang2021grasp} is a generic framework for healthcare models, which leverages the information extracted from patients with similar conditions to enhance the cohort representation learning results.
    \item M3Care~\cite{zhang2022m3care} resolves the missing modality issue in EHR data by utilizing similar patients' existing modalities. However, it does not address the issue of missing features within available modalities.
    \item SAFARI~\cite{ma2022safari} learns patient health representations by applying a clinical-fact-inspired, task-agnostic correlational sparsity prior to medical feature correlations, using a bi-level optimization process that involves both inter- and intra-group correlations.
    \item AICare~\cite{ma2023aicare} also includes a multi-channel feature extraction module and an adaptive feature importance recalibration module. It learns personalized health status embeddings with static and dynamic features.
\end{itemize}

\subsubsection{\textbf{Ablation Models}}
\mbox{}\\
Ablation models include \modelname{}$_{\text{-proto.}}$ and \modelname{}$_{\text{-calib.}}$.
\begin{itemize}[leftmargin=*]
    \item \modelname{}$_{\text{-proto.}}$ removes the confidence-aware prototype similar patient learner and reserves the feature-missing-aware calibration process. We apply the patients hidden representation \( \bm{z} \) for downstream GRU module.
    \item \modelname{}$_{\text{-calib.}}$ removes the feature confidence learner from the missing-feature-aware calibration process. As the confidence-aware prototype patient learner requires the feature confidence \( \bm{C} \) as input, we set the \( \bm{C} = 1 \) for each feature.
\end{itemize}

\subsection{Implementation Details}

\subsubsection{\textbf{Hardware and Software Configuration}}
\mbox{}\\
All runs are trained on a single Nvidia RTX 3090 GPU with CUDA 11.8. The server's system memory (RAM) size is 64GB. We implement the model in Python 3.11.4, PyTorch 2.0.1~\cite{paszke2019pytorch}, PyTorch Lightning 2.0.5~\cite{falcon2019lightning}, and pyehr~\cite{zhu2024pyehr}.

\subsubsection{\textbf{Model Training and Hyperparameters}}
\mbox{}\\
AdamW~\cite{loshchilov2017decoupled} is employed with a batch size of 1024 patients. All models are trained for 50 epochs with an early stopping strategy based on AUPRC after 10 epochs without improvement. The learning rate ${0.01,0.001,0.0001}$ and hidden dimensions ${64, 128}$ are tuned using a grid search strategy on the validation set. The searched hyperparameter for \modelname{} is: 128 hidden dimensions, 0.001 learning rate, and 256 prototype patient cluster numbers. Performance is reported in the form of mean±std of 5 runs with random seeds ${0,1,2,3,4}$ for MIMIC-III and MIMIC-IV datasets, and apply bootstrapping on all test set samples 10 times for the Challenge-2012 and eICU datasets, following practices in \citet{ma2023aicare}.

\begin{table*}[!ht]
        \tiny
        \centering
        \caption{\textit{Benchmarking results on MIMIC-III, MIMIC-IV, Challenge-2012, and eICU datasets.} \textbf{Bold} indicates the best performance. Performance is reported in the form of mean±std. All metric scores are multiplied by 100 for readability purposes.}
        \label{tab:motality_readmission_prediction_exp}
\resizebox{\textwidth}{!}{
    \begin{tabular}{c|cc|cc|cc|cc|cc|cc}
\toprule
\textbf{Dataset}     & \multicolumn{2}{c|}{\textbf{MIMIC-III Mortality}} & \multicolumn{2}{c|}{\textbf{MIMIC-III Readmission}} & \multicolumn{2}{c|}{\textbf{MIMIC-IV Mortality}} & \multicolumn{2}{c|}{\textbf{MIMIC-IV Readmission}} & \multicolumn{2}{c|}{\textbf{Challenge-2012 Mortality}} & \multicolumn{2}{c}{\textbf{eICU Mortality}} \\ \midrule
\textbf{Metric}      & AUPRC($\uparrow$)   & AUROC($\uparrow$) & AUPRC($\uparrow$)   & AUROC($\uparrow$) & AUPRC($\uparrow$)   & AUROC($\uparrow$)  & AUPRC($\uparrow$)    & AUROC($\uparrow$)   & AUPRC($\uparrow$)  & AUROC($\uparrow$)  & AUPRC($\uparrow$)   & AUROC($\uparrow$)  \\ \midrule
MICE                   & 52.40±0.57                           & 85.43±0.23                           & 50.61±0.47                           & 77.97±0.60                           & 49.89±0.78                           & 84.37±0.13                           & \multicolumn{1}{c|}{75.84±0.25}                           & 45.28±0.38                           & 32.26±2.85                           & 72.41±1.46                           & 44.44±4.16                           & 85.19±2.97                           \\
GRU-D                  & 45.31±3.22                           & 81.72±2.46                           & 42.13±3.19                           & 73.31±1.65                           & 48.79±1.57                           & 85.02±0.50                           & \multicolumn{1}{c|}{76.47±0.53}                           & 45.03±0.71                           & 25.43±3.27                           & 63.75±1.76                           & 42.59±3.33                           & 83.85±2.31                           \\
RETAIN                 & 51.76±0.86                           & 85.57±0.43                           & 47.53±0.48                           & 77.42±0.38                           & 54.06±0.71                           & 86.24±0.36                           & \multicolumn{1}{c|}{78.54±0.38}                           & 49.93±0.73                           & 30.23±2.24                           & 69.82±1.89                           & 39.89±3.08                           & 82.53±2.45                           \\
AdaCare                & 52.28±0.50                           & 85.73±0.19                           & 48.76±0.35                           & 77.65±0.32                           & 50.45±0.80                           & 83.96±0.13                           & \multicolumn{1}{c|}{77.00±0.20}                           & 48.57±0.29                           & 33.10±3.42                           & 69.66±1.57                           & 42.48±3.61                           & 83.91±2.32                           \\
ConCare                & 51.45±0.76                           & 86.18±0.14                           & 47.45±0.96                           & 77.74±0.32                           & 49.97±1.08                           & 85.41±0.40                           & \multicolumn{1}{c|}{77.47±0.19}                           & 47.17±0.84                           & 30.24±3.00                           & 70.19±2.54                           & 44.40±4.35                           & 85.05±2.93                           \\
GRASP                  & 53.59±0.33                           & 86.54±0.17                           & 50.21±0.22                           & 78.14±0.35                           & 54.41±0.46                           & 86.08±0.17                           & \multicolumn{1}{c|}{78.50±0.22}                           & 50.22±0.26                           & 26.03±3.43                           & 67.14±2.61                           & 45.41±4.05                           & 85.69±2.38                           \\
M3Care                 & 51.68±1.03                           & 86.23±0.42                           & 49.00±0.71                           & 78.00±0.55                           & 52.95±0.71                           & 84.90±0.37                           & \multicolumn{1}{c|}{77.31±0.42}                           & 49.22±0.69                           & 32.63±2.54                           & 73.26±1.67                           & 44.95±4.32                           & 85.44±2.56                           \\
SAFARI                 & 45.92±1.01                           & 85.10±0.24                           & 45.59±0.35                           & 77.01±0.21                           & 46.58±0.55                           & 46.58±0.55                           & \multicolumn{1}{c|}{76.05±0.38}                           & 44.78±0.69                           & 28.94±3.16                           & 70.67±1.76                           & 35.26±3.59                           & 80.10±2.40                           \\
AICare                 & 51.37±0.70                           & 85.40±0.48                           & 47.06±1.16                           & 76.23±0.84                           & 49.76±0.86                           & 84.62±0.28                           & \multicolumn{1}{c|}{76.07±0.43}                           & 45.88±1.12                           & 23.99±2.48                           & 67.35±2.20                           & 42.80±3.79                           & 84.26±2.64                           \\
\midrule
\modelname{}$_\text{-proto.}$ & 55.52±0.34                           & 87.28±0.11                           & 51.17±0.25                           & 78.66±0.22                           & 55.76±0.90                           & \textbf{86.82±0.16} & \multicolumn{1}{c|}{79.12±0.44}                           & 50.75±0.65                           & 30.92±2.95                 & 68.87±2.61                & {50.03±3.97}                 & {\textbf{85.93±1.67}} \\
\modelname{}$_\text{-calib.}$ & 56.16±0.42                           & 87.33±0.22                           & 49.13±2.11                           & 77.87±0.88                           & 55.18±0.77                           & 86.57±0.20                           & \multicolumn{1}{c|}{78.66±0.45}                           & 50.62±0.70                           & {30.42±2.96}                 & {71.07±2.13}                 & {46.92±3.20}                 & {84.69±1.39} \\
\midrule
\modelname{}                  & \textbf{57.02±0.38} & \textbf{87.34±0.22} & \textbf{51.31±1.02} & \textbf{78.76±0.59} & \textbf{55.92±0.75} & \textbf{86.82±0.20} & \multicolumn{1}{c|}{\textbf{79.14±0.33}} & \textbf{51.04±0.70} & \textbf{33.60±3.41} & \textbf{73.47±1.11} & \textbf{50.41±3.63} & 85.82±1.43 \\
\bottomrule
\end{tabular}
}
\end{table*}

\section{Experimental Results and Analysis}\label{sec:experiments}

We conduct the in-hospital mortality and 30-day readmission prediction task on MIMIC-III and MIMIC-IV datasets, in-hospital mortality prediction task on Challenge-2012 and eICU datasets. 

\subsection{Experimental Results}

Table~\ref{tab:motality_readmission_prediction_exp} depicts the performance evaluation of baseline methods, \modelname{}, and its reduced versions for ablation study on four datasets under two prediction tasks. Additionally, we conduct t-test based on the AUPRC metric, the \modelname{}'s performance improvement against all models are all statistically significant with p-value $<$ 0.01, which underscores that \modelname{} significantly outperforms all baseline models. Specifically, \modelname{} outperforms models focused solely on enhancing feature representations with attention mechanisms (like RETAIN, AdaCare, ConCare, AICare, SAFARI), by integrating missing feature status into these mechanisms for improved attention calibration and feature representation. It also exceeds models using similar patient information (such as GRASP, M3Care), showing the value of missing feature status in refining prototype patient representations for better performance. \modelname{}'s advantage over GRU-D, which only considers local patient visit-based feature missing status, highlights the significance of a global perspective on overall feature missing rates for effective feature representation across patients.

\subsection{Ablation Study}

\subsubsection{\textbf{Comparing with Reduced Versions}}
\mbox{}\\
\modelname{} outperforms \modelname{}$_{\text{-proto.}}$ and \modelname{}$_{\text{-calib.}}$ on main metric AUPRC. This indicates that the two designed learners can enhance patient feature representations from different perspectives: the patient's individual health data utilized by the feature confidence learner based on the attention mechanism and the prototype similar patient representations utilized by the prototype similar patient learner.

\subsubsection{\textbf{Comparing with Internal Components}}
\mbox{}\\
To deeply explore the impact of components within each module, we conduct experiments in Table~\ref{tab:ablation_exp}, showing \modelname{} outperforms all baselines. The symbol \( \phi \) denotes similarity measure, detailed in Equation~\ref{eq:phi}. The term \( \bm{z} \) alone indicates the use of L2 distance for patient similarity measure, whereas \( \bm{z}, \bm{C} \) additionally incorporates feature confidence, enhancing the model's discriminative capability. Comparing the roles of various components within the feature confidence learner, the performance when considering both global feature missing rate $\rho$ and local patient's time interval $\tau$ is higher than considering any single component, which illustrates the necessity of considering the feature missing status from both global and local perspectives. When only considering the local perspective, its performance actually worsens, which is consistent with our observation of the performance of GRU-D. As for confidence-aware prototype patient learner, the performance of confidence-aware patient similarity measurement surpasses that without considering feature confidence, which also shows the impact of missing feature status on measuring similar patients.

\begin{table}[!ht]
        \footnotesize
        \centering
        \caption{\textit{Performance comparison of internal components on the MIMIC-IV mortality prediction task.} ``Feat. Conf.'' means ``Feature Confidence'' and ``Sim. Meas.'' denotes ``Similarity Measure''. \textbf{Bold} denotes the best performance within each components, \textbf{\textcolor{red}{Red}} denotes the highest performance among all comparisons. Performance is reported in the form of mean±std. All metric scores are multiplied by 100 for readability purposes.}
        \label{tab:ablation_exp}
\begin{tabular}{c|ccc|cc}
\toprule
\multirow{2}{*}{\textbf{Comparisons}}        & \multicolumn{3}{c|}{\textbf{Components}}                & \multicolumn{2}{c}{\textbf{Metrics}}           \\ \cline{2-6} 
                                    & +$\rho$ & +$\tau$ & +$\phi$                    & AUPRC($\uparrow$) & AUROC($\uparrow$) \\ \midrule
\multirow{4}{*}{Feat. Conf.} & /       & /       & /                          & 54.42±0.43        & 86.22±0.30        \\
                                    & $\surd$ & /       & /                          & 55.11±0.53        & 86.56±0.33        \\
                                    & /       & $\surd$ & /                          & 52.67±2.52        & 86.43±0.55        \\
                                    & $\surd$ & $\surd$ & /                          & \textbf{55.76±0.90}        & \textbf{86.82±0.16}        \\ \midrule
\multirow{2}{*}{Sim. Meas.} & /       & /       & \( \bm{z} \)               & 55.18±0.77        & 86.57±0.20        \\
                                    & /       & /       & \( \bm{z} \), \( \bm{C} \) & \textbf{55.25±0.75}        & \textbf{86.60±0.19}        \\ \midrule
\modelname{}                               & $\surd$ & $\surd$ & \( \bm{z} \), \( \bm{C} \) & \textbf{\textcolor{red}{55.92±0.75}}        & \textbf{\textcolor{red}{86.82±0.20}}        \\
\bottomrule
\end{tabular}
\end{table}

\subsection{Observations and Analysis}

\subsubsection{\textbf{Robustness to Data Sparsity}}
\begin{figure}[!ht]
\centering
\subfigure[Volume-level Sparsity]{
  \includegraphics[width=0.45\linewidth]{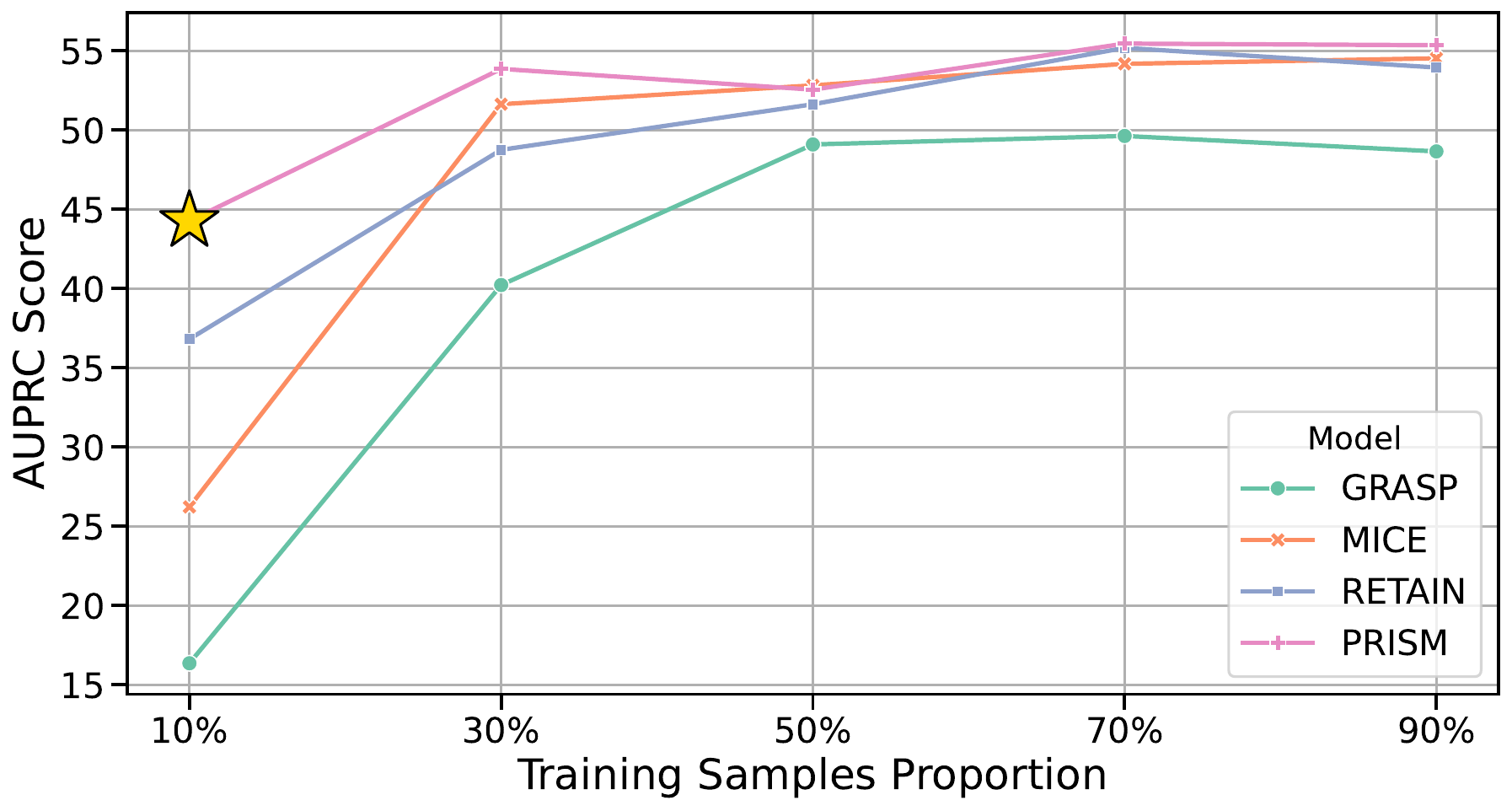}
  \label{fig:volume_rates}
}
\subfigure[Feature-level Sparsity]{
  \includegraphics[width=0.45\linewidth]{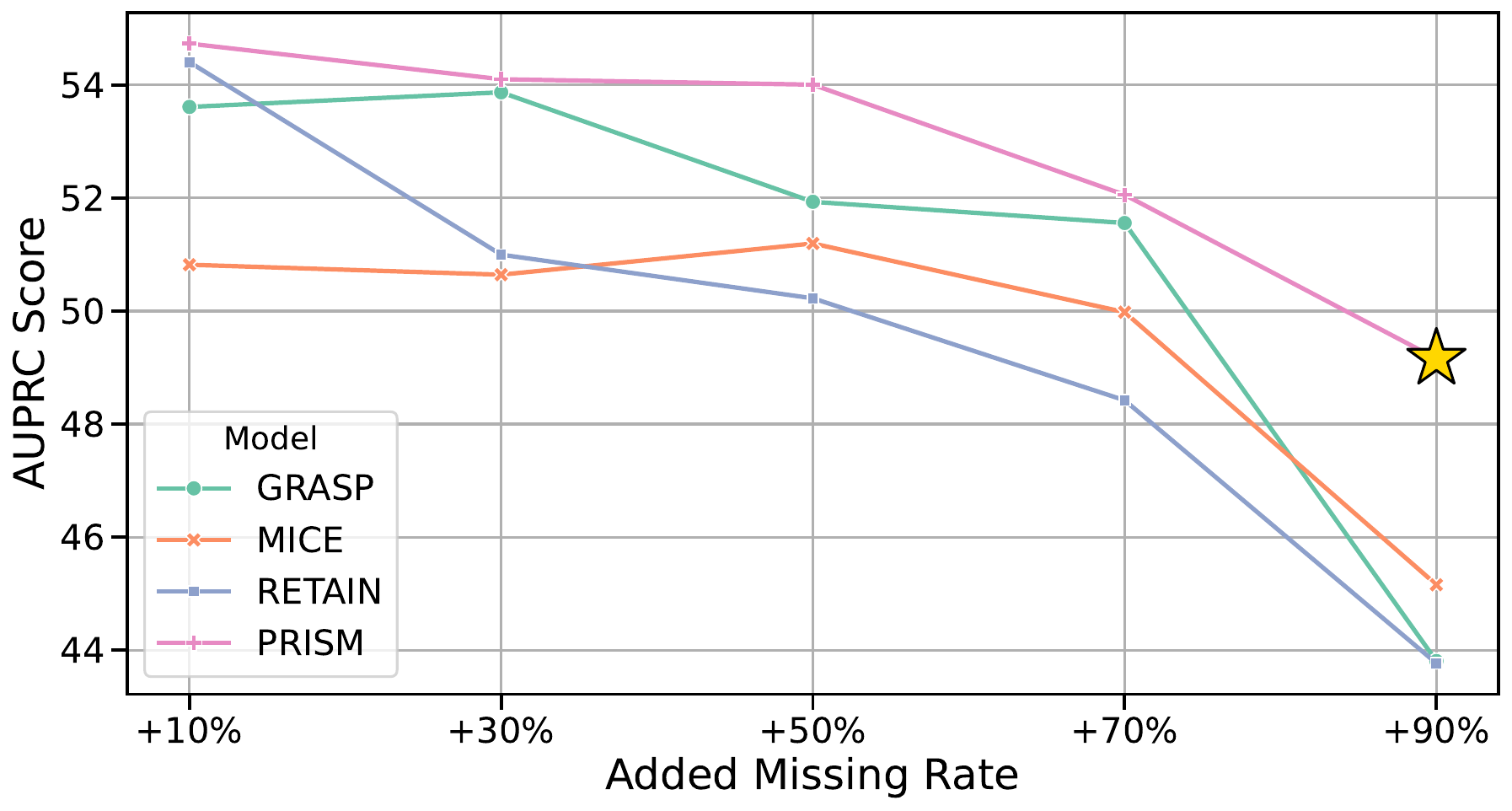}
  \label{fig:missing_rates}
}
\caption{\textit{AUPRC performance across 5 sparsity levels in MIMIC-IV in-hospital mortality prediction task.} \modelname{} significantly outperforms other models in extremely sparse scenarios on both sparsity settings.}
\label{fig:sparsity}
\end{figure}

\mbox{}\\
To assess \modelname{}'s performance under conditions of data sparsity, we compare it with leading models such as GRASP, MICE(LSTM), and RETAIN in MIMIC-IV in-hospital mortality prediction task. At the volume-level in Figure~\ref{fig:volume_rates}, we reduce the data samples in training set, while at the feature-level in Figure~\ref{fig:missing_rates}, we intentionally increase the missing feature rates beyond the original missing rate. \modelname{} excels in both settings. Notably, in situations of extreme data sparsity, such as using only 10\% training data and with an overall 97.47\% feature missing rate, \modelname{} significantly outperforms the other models, highlighting its robustness in handling sparse data.

\subsubsection{\textbf{Sensitiveness to Cohort Size and Effectiveness of Missing-Feature-Aware Module}}
\begin{figure}[!ht]
  \centering
\subfigure[AUPRC]{
  \includegraphics[width=0.45\linewidth]{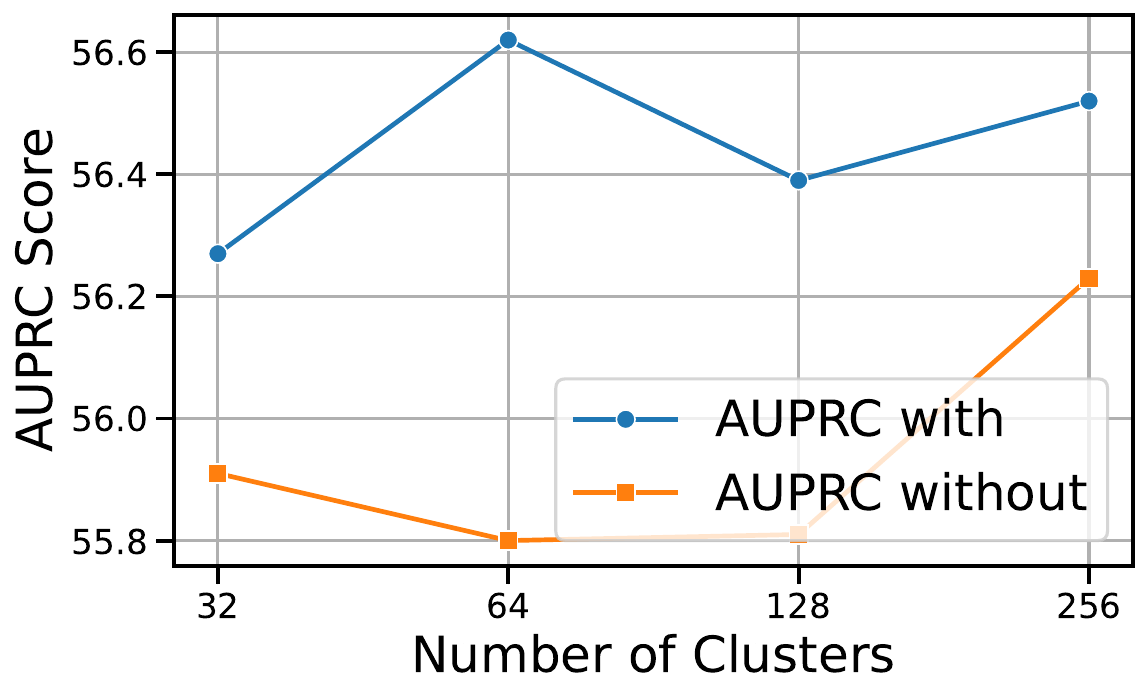}
  \label{fig:clusters_auprc_scores}
}
\subfigure[F1]{
  \includegraphics[width=0.45\linewidth]{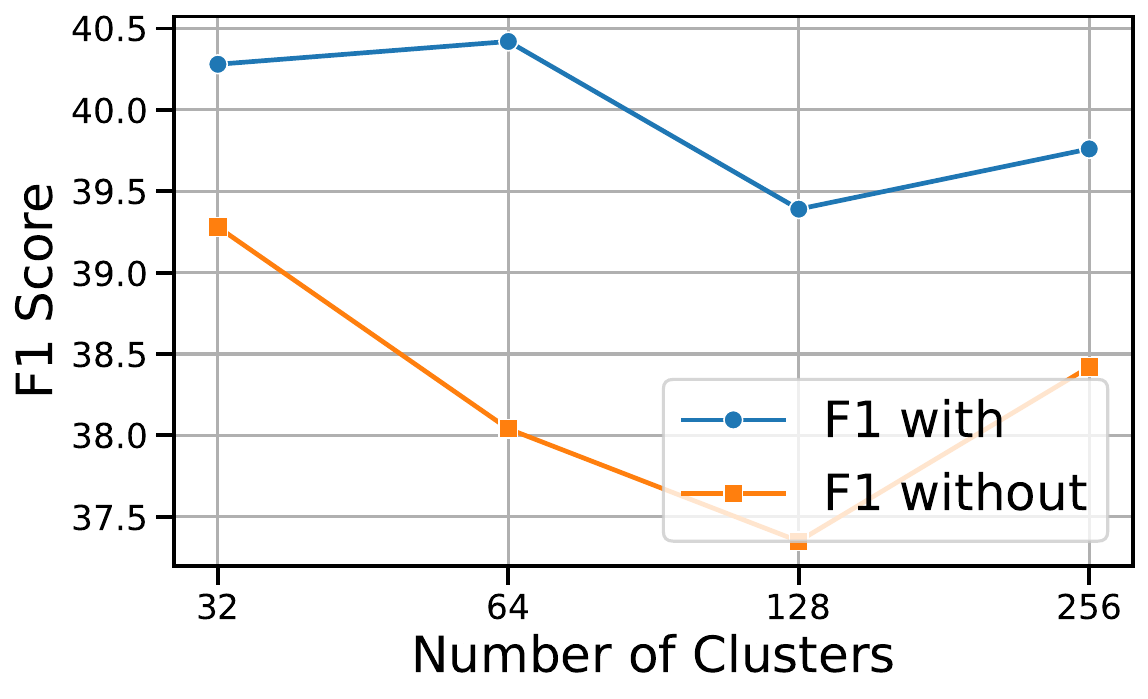}
  \label{fig:clusters_f1_scores}
}
  \caption{\textit{AUPRC and F1 score performance on various prototype patient cohort size in MIMIC-IV in-hospital mortality prediction task.} With the missing-feature-aware module, \modelname{} outperforms its counterpart without. It also shows \modelname{} is not sensitive to the cohort size.}
  \label{fig:diverse_prototype_clusters}
\end{figure}

\mbox{}\\
We conduct a detailed analysis to examine the impact of prototype patients cohort diversity and the role of a missing-feature-aware module in patient prototypes. Figure~\ref{fig:diverse_prototype_clusters} shows that integrating a missing-feature-aware module consistently enhances performance across various cluster sizes, as indicated by the superior AUPRC and F1 score. Furthermore, the relatively consistent performance across different cluster sizes demonstrates that our model is not overly sensitive to the number of clusters, highlighting its adaptability and robustness in managing various cohort sizes.

\subsubsection{\textbf{Variations of Similarity Measures}}
\mbox{}\\
We assess the performance of \modelname{} by comparing it against standard similarity metrics commonly used in evaluating patient similarities. Table~\ref{tab:variation_similarity_metrics} details the results of this comparison on the MIMIC-IV in-hospital mortality prediction task. As demonstrated in the table, \modelname{}, leveraging a confidence-aware patient similarity measure, consistently surpasses traditional metrics such as cosine similarity, and L1 and L2 distances. This showcases the effectiveness of \modelname{}'s similarity metric in measuring the impact of missing features.

\begin{table}[!ht]
\footnotesize
\centering
\caption{\textit{Comparing different similarity measures in MIMIC-IV in-hospital mortality prediction task.} Performance is reported in the form of mean±std. All metric scores are multiplied by 100 for readability purposes.}
\label{tab:variation_similarity_metrics}
\begin{tabular}{c|c|c|c}
\toprule
\textbf{Measure} & \textbf{AUPRC}($\uparrow$) & \textbf{AUROC}($\uparrow$) & \textbf{F1}($\uparrow$) \\ 
\midrule
Cosine & 55.58±0.43 & 86.81±0.10 &  45.44±1.99 \\
L1     & 55.59±0.54 & 86.73±0.20 &  45.20±1.91 \\
L2     & 55.38±0.51 & 86.76±0.09 &  43.69±1.25 \\
\midrule
\modelname{}   & \textbf{55.92±0.75} & \textbf{86.82±0.20} & \textbf{45.47±2.26} \\
\bottomrule
\end{tabular}
\end{table}

\subsubsection{\textbf{Model Efficiency and Complexity}}
\mbox{}\\

We evaluate the efficiency and complexity of \modelname{} in terms of parameter count, runtime, and data preparation time. \modelname{} achieves a competitive balance between a low parameter count (215K) and an efficient runtime (69.03s for 5 epochs) on the MIMIC-IV dataset, using a hidden dimension of 128 and batch size of 1024. Table~\ref{tab:data_preparation_time} presents the data preparation time for \modelname{} on MIMIC-III, MIMIC-IV, Challenge-2012, and eICU datasets, including preprocessing, training, validation, and testing. The \modelname{} pipeline seamlessly computes feature missing statuses in the LOCF pipeline with little extra cost. Moreover, the entire pipeline can be completed within 10 minutes for each dataset.

\begin{table}[!ht]
        \centering
        \caption{\textit{Data preparation time comparison.} Prep. denotes Preprocessing, Val. denotes Validation. All with 50 epochs of training, validate at the end of each epoch. The units are seconds.}
        \label{tab:data_preparation_time}
\resizebox{\columnwidth}{!}{
    \begin{tabular}{c|ccc|ccc}
\toprule
\textbf{Dataset}     & \multicolumn{3}{c|}{\textbf{MIMIC-III}} & \multicolumn{3}{c}{\textbf{MIMIC-IV}} \\ \midrule
Prep.      &  w/o Impute & LOCF   & +Ours  & w/o Impute & LOCF   & +Ours  \\ \midrule
Time & 77.06 & 300.63 & 432.19 & 104.67 & 358.29 & 502.68 \\ \midrule
Pipeline      &  Data Prep.   & Train+Val. & Test & Data Prep.   & Train+Val. & Test  \\ \midrule
Time & 432.19 & 505.83 & 10.25 & 502.68 & 510.73 & 12.39 \\
\toprule
\textbf{Dataset}     & \multicolumn{3}{c|}{\textbf{Challenge-2012}} & \multicolumn{3}{c}{\textbf{eICU}} \\ \midrule
Prep.    & w/o Impute & LOCF   & +Ours  & w/o Impute & LOCF   & +Ours  \\ \midrule
Time     & 18.46      & 56.59  & 85.14  & 52.99 & 146.54 & 170.21 \\ \midrule
Pipeline &  Data Prep.   & Train+Val. & Test & Data Prep.   & Train+Val. & Test  \\ \midrule
Time     & 85.14 & 85.55 & 1.47 & 170.21 & 268.17 & 4.99 \\
\bottomrule
\end{tabular}
}
\end{table}

\subsubsection{\textbf{Cross-Feature Attention Map}}
\begin{figure}[!ht]
    \centering
    \includegraphics[width=1.0\linewidth]{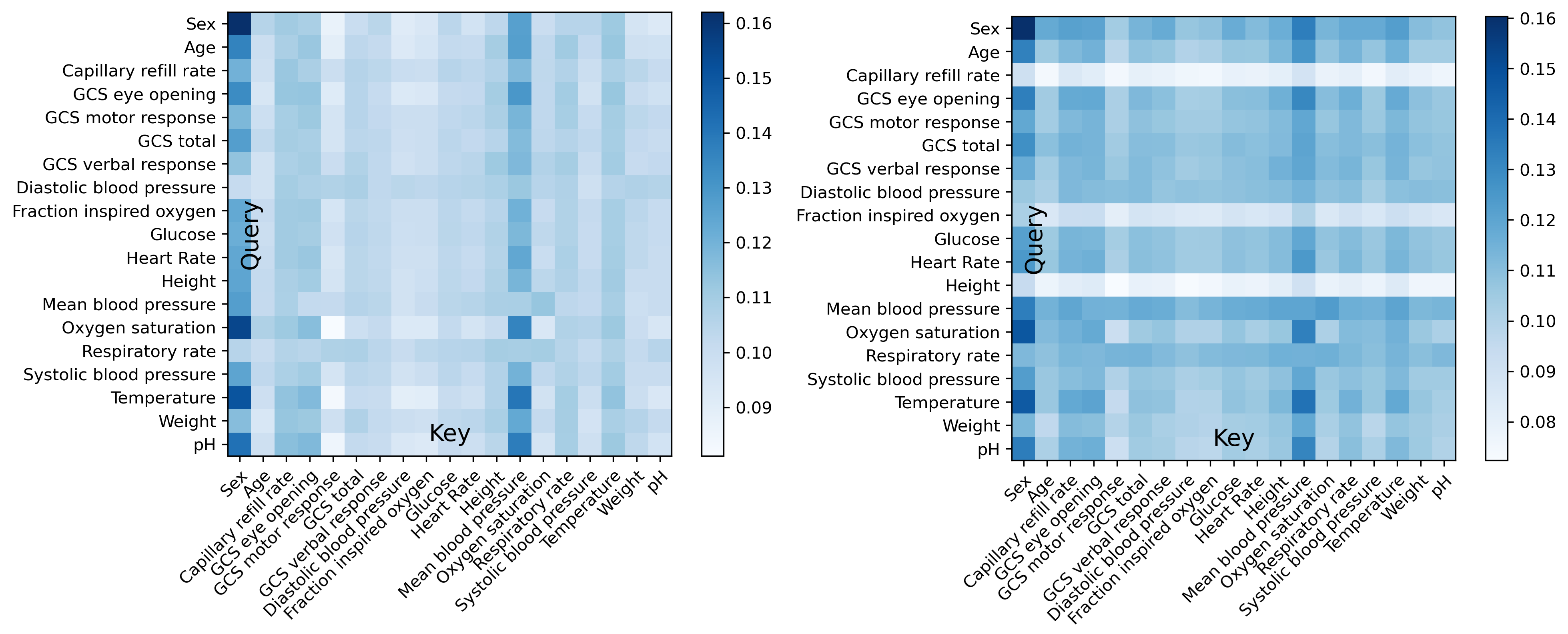}
    \caption{\textit{Cross-feature attention maps from \modelname{}: Without (Left) / With (Right) feature confidence calibration.} The maps use data from a single MIMIC-IV patient record to show \modelname{}'s reduction in attention to unreliable features (capillary refill rate, fraction inspired oxygen, height) due to high missing rates.}
    \label{fig:cross_feature_attention}
\end{figure}

\mbox{}\\
Figure~\ref{fig:cross_feature_attention} presents the cross-feature attention map from \modelname{}, contrasting average attention weights with and without feature confidence calibration process. This visualization, based on a single diagnostic record from randomly selected MIMIC-IV patients, plots Key features on the x-axis against Query features on the y-axis. Notably, \modelname{} reduces attention on features like capillary refill rate, fraction inspired oxygen, and height, which have high missing rates (99.64\%, 93.62\%, and 99.61\%, respectively) in the dataset. 
\modelname{} accurately recognizes and calibrates the features with high missing rates, thereby causing the three horizontal lines on the right side of the graph to appear distinctively whiter, showcasing attention-based feature learner module's interpretability which other baselines lack.

\subsubsection{\textbf{Patient Representation Visualization}}
\begin{figure}[!ht]
  \centering
\subfigure[Without calibration]{
  \includegraphics[width=0.45\linewidth]{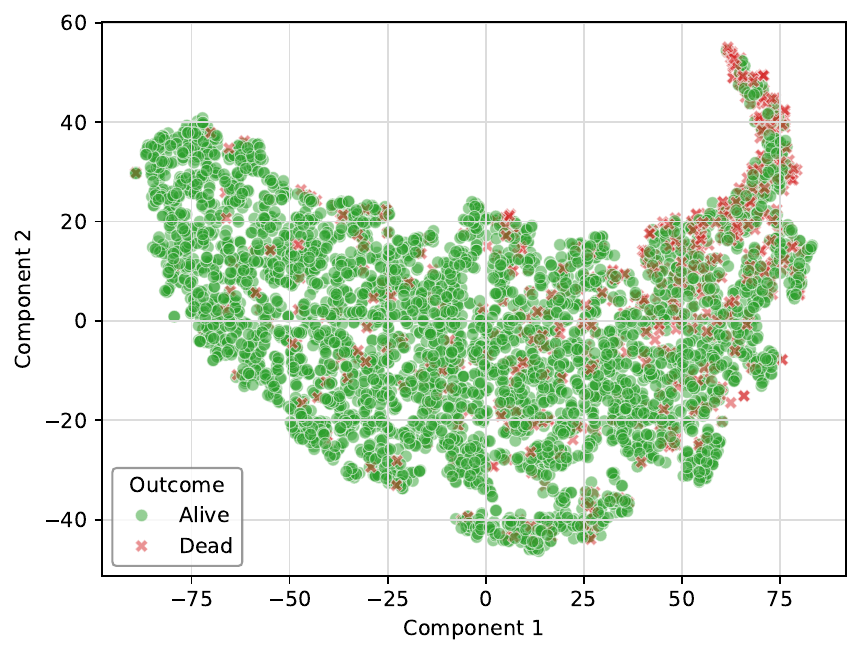}
  \label{fig:embedding_ablation}
}
\subfigure[With calibration]{
  \includegraphics[width=0.45\linewidth]{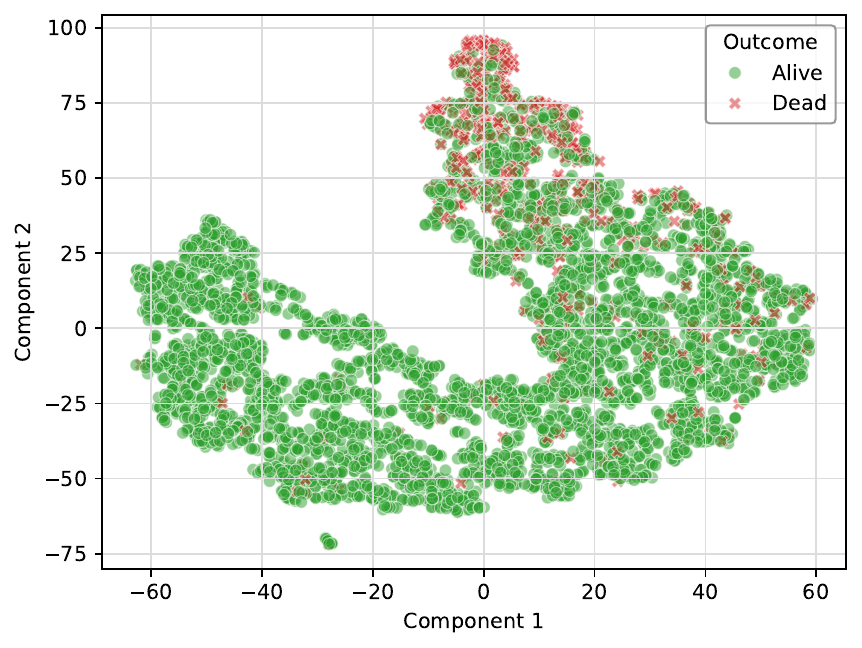}
  \label{fig:embedding_prism}
}
\caption{\textit{t-SNE visualization of patient representations from \modelname{}.} (a) shows the embeddings without the missing-feature-calibration process, and (b) depicts embeddings with the process. (b)'s representations are more compact among dead outcome patients, showcasing it learns better representations. }
  \label{fig:embedding_visualization}
\end{figure}

\mbox{}\\
To investigate the impact of the missing-feature-calibration process on the hidden representations of patient data, we apply t-SNE to project these representations onto a two-dimensional space using the test set of all patients from the MIMIC-IV dataset. Figure~\ref{fig:embedding_visualization} illustrates the t-SNE embeddings of the patient representations generated by \modelname{}, both with and without the application of the missing-feature-calibration process. The calibrated representations (Figure~\ref{fig:embedding_prism}) exhibit improved separation and compactness, particularly for patients with mortal outcomes, compared to the representations without calibration (Figure~\ref{fig:embedding_ablation}). This observation suggests that the missing-feature-calibration process enables \modelname{} to learn more informative and discriminative representations of patient data by effectively handling missing features and capturing the underlying patterns and relationships within the data.

\subsubsection{\textbf{Feature Decay Rates Observation}}
\begin{figure}[!ht]
    \centering
    \includegraphics[width=0.95\linewidth]{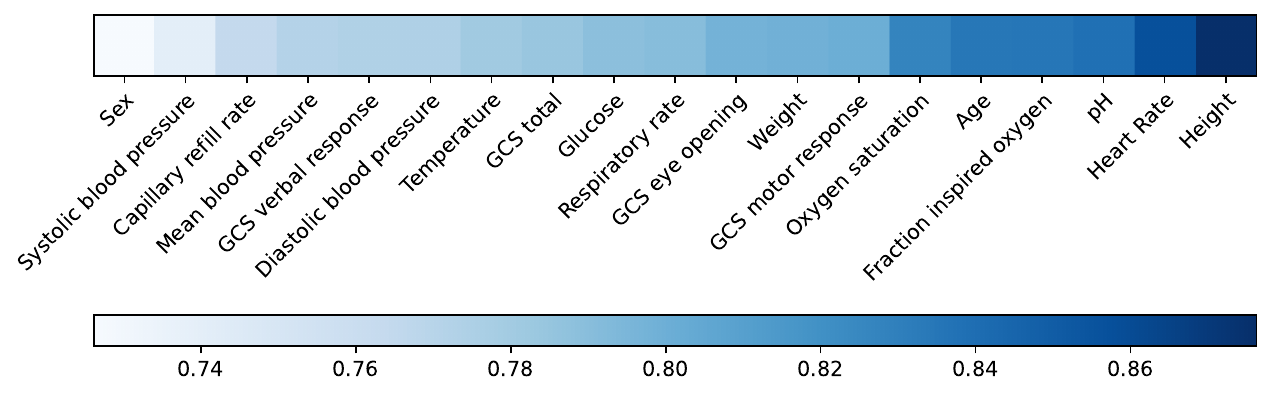}
    \caption{\textit{Adaptive learning feature decay rates.} The graph shows varying decay rates: high for acute-indicator features like heart rate and pH, and low for longer-term relevant features like sex and systolic blood pressure.}
    \label{fig:feature_decay_rates}
\end{figure}

\mbox{}\\
Figure~\ref{fig:feature_decay_rates} displays the decay rates of adaptive learning for various features, indicating how their importance diminishes over time. Higher decay rates suggest that the model prioritizes immediate changes in features like heart rate and pH, which is crucial for detecting acute medical conditions such as shock or infection. Conversely, features like sex and systolic blood pressure exhibit lower decay rates, highlighting their relevance in long-term analysis. Notice that the feature `Height' exhibits short-term dynamics, likely due to being absent in over 99\% of the data. Consequently, the rapid decay of the `Height' feature, resulting from its high missing rate, does not significantly influence clinical decisions over the long term, which aligns with our intuition.

\section{Limitations and Further Work}

We identify key limitations and future research directions:
\begin{itemize}[leftmargin=*]
    \item \textbf{Fairness Concerns:} Evaluate the model's fairness across various demographic groups and explore bias in similar patient cohorts.
    \item \textbf{Scalability Issues:} Assess the scalability of the proposed model to larger datasets or its integration within real-time healthcare systems.
    \item \textbf{Prototype Patient Representations:} Understand the diversity of prototype patient representations and explore more intricate mechanisms for prototype generation beyond similarity metrics.
\end{itemize}

\section{Conclusions}

In this work, we propose \modelname{}, a prototype patient representation learning framework to address the sparsity issue of EHR data. \modelname{} perceives and calibrates for missing features, thereby refining patient representations via a confidence-aware prototype patient learner. Significant performance improvements and detailed experimental analysis on four real-world datasets' in-hospital mortality and 30-day readmission prediction tasks show \modelname{}'s effectiveness. The work marks a crucial step towards more reliable and effective utilization of EHR data in healthcare, offering a potent solution to the prevalent issue of data sparsity in clinical decision-making.

\section*{Ethical Statement}
This study analyzes de-identified EHR data from MIMIC-III, MIMIC-IV, PhysioNet Challenge 2012, and eICU datasets. We adhere to data use agreements, prioritize patient privacy, and strive for unbiased, equitable findings that reflect the complexity of medical data. Our methodology aims to minimize potential harm and uphold ethical research standards.

\begin{acks}
This work was supported by the National Science and Technology Major Project(2022ZD0116401), the National Natural Science Foundation of China (U23A20468), and Xuzhou Scientific Technological Projects (KC23143).
\end{acks}

\clearpage
\newpage
\balance
\bibliographystyle{ACM-Reference-Format}
\bibliography{ref}

\end{document}